\documentclass[conference,A4]{IEEEtran}
\usepackage{pifont}
\newcommand{\cmark}{\ding{51}}%
\newcommand{\xmark}{\ding{53}}%
\IEEEoverridecommandlockouts
\usepackage{cite}
\usepackage{amsmath,amssymb,amsfonts}
\usepackage{algorithmic}
\usepackage{graphicx}
\usepackage{textcomp}
\usepackage{xcolor}
\usepackage{multirow}
\usepackage{hyperref}
\def\BibTeX{{\rm B\kern-.05em{\sc i\kern-.025em b}\kern-.08em
    T\kern-.1667em\lower.7ex\hbox{E}\kern-.125emX}}
\makeatletter
\let\old@ps@headings\ps@headings
\let\old@ps@IEEEtitlepagestyle\ps@IEEEtitlepagestyle
\def\confheader#1{%
\def\ps@IEEEtitlepagestyle{
\old@ps@IEEEtitlepagestyle
\def\@oddhead{\strut\hfill#1\hfill\strut}
\def\@evenhead{\strut\hfill#1\hfill\strut}
}
\ps@headings
}
\makeatother
    
\begin{document}

\makeatletter 
\newcommand{\linebreakand}{%
  \end{@IEEEauthorhalign}
  \hfill\mbox{}\par
  \mbox{}\hfill\begin{@IEEEauthorhalign}
}
\makeatother

\title{AGILE: Approach-based Grasp Inference Learned from Element Decomposition}

\author{\IEEEauthorblockN{MohammadHossein Koosheshi\IEEEauthorrefmark{1}\IEEEauthorrefmark{2},
Hamed Hosseini\IEEEauthorrefmark{1},
Mehdi Tale Masouleh\IEEEauthorrefmark{1}, \\
Ahmad Kalhor\IEEEauthorrefmark{1}, and
Mohammad Reza Hairi Yazdi\IEEEauthorrefmark{2}}
\IEEEauthorblockA{\IEEEauthorrefmark{1}Human and Robot Interaction  Lab, School of Electrical and Computer Engineering,  \\ 
University of Tehran, Tehran, Iran}
\IEEEauthorblockA{\IEEEauthorrefmark{2} School of Mechanical Engineering,  \\ 
University of Tehran, Tehran, Iran}}


\maketitle

\begin{abstract}
Humans, this species expert in grasp detection, can grasp objects by taking into account hand-object positioning information. This work proposes a method to enable a robot manipulator to learn the same, grasping objects in the most optimal way according to how the gripper has approached the object. Built on deep learning, the proposed method consists of two main stages. In order to generalize the network on unseen objects, the proposed Approach-based Grasping Inference involves an element decomposition stage to split an object into its main parts, each with one or more annotated grasps for a particular approach of the gripper. Subsequently, a grasp detection network utilizes the decomposed elements by Mask R-CNN and the information on the approach of the gripper in order to detect the element the gripper has approached and the most optimal grasp. In order to train the networks, the study introduces a robotic grasping dataset collected in the Coppeliasim simulation environment. The dataset involves 10 different objects with annotated element decomposition masks and grasp rectangles. The proposed method acquires a 90\% grasp success rate on seen objects and 78\% on unseen objects in the Coppeliasim simulation environment. Lastly, simulation-to-reality domain adaptation is performed by applying transformations on the training set collected in simulation and augmenting the dataset, which results in a 70\% physical grasp success performance using a Delta parallel robot and a 2-fingered gripper.
\end{abstract}

\begin{IEEEkeywords}
Deep Learning, Robotic Grasping, Grasping Dataset, Delta Parallel Robot, Domain Adaptation.
\end{IEEEkeywords}

\section{Introduction}
Robotic grasping is one of the most important yet complicated robotic tasks. Detecting the grasp points of an object is a routine for a human. But in addition to their abilities to grasp complex objects, humans can process the information of the spatial relationship between their hand and the object to select the best grasping pose among the possible ones. Furthermore, to grasp objects that are geometrically new to the ones seen before, a human finds the grasp candidate parts by breaking the complex objects into their elements, for which the best grasp pose is inherently obvious. To take a step further, humans can combine the hand-object and element decomposition information to predict the most optimal grasp based on the approach of their hand toward a specific part of the object. This work focuses on training a robotic manipulator to process the same information for a successful and optimized grasp. The proposed Approach-based Grasp Inference Learned from Element Decomposition (AGILE) consists of two main stages, the element decomposition and grasp detection stages. Therefore, in the following paragraphs, the relevant literature on robotic grasp detection and object element decomposition is analyzed to provide a framework for the subsequent discussion.

\begin{figure*}[!t]
\centerline{\includegraphics[ width = 1\textwidth]{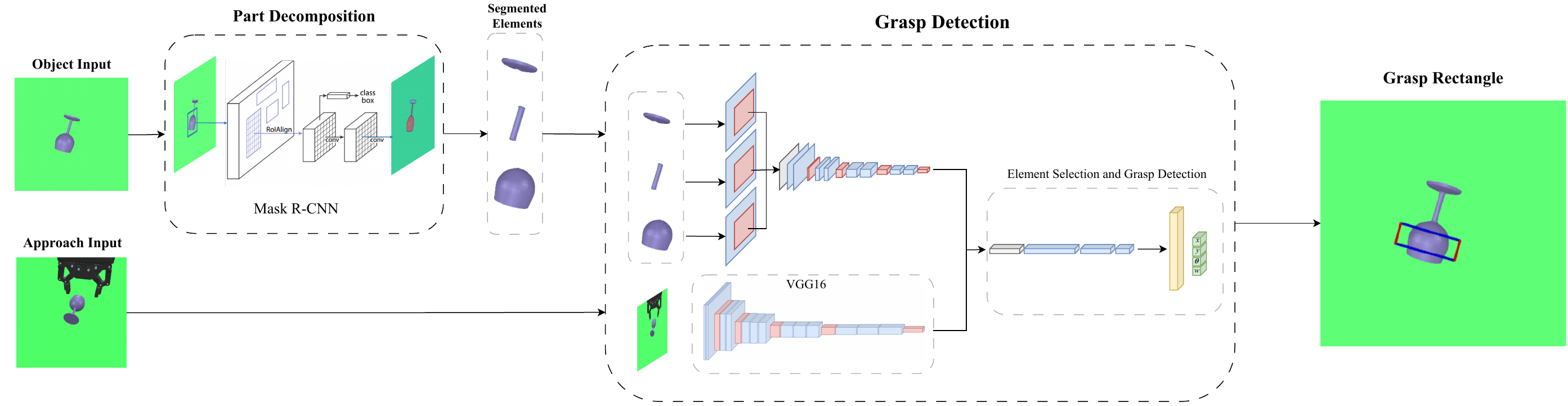}}
\caption{Overview of the proposed Approach-based Grasp Inference method by decomposing the elements of the object. Starting from the top view of the object, Mask R-CNN decomposes the object into its primitive elements. The decomposed parts are fed into a grasp detection network, which decides upon the best graspable part by also consuming the information on the approach of the gripper and detecting the most optimal grasp based on the presented approach information.}
\label{fig:network}
\end{figure*}

\textbf{Robotic Grasp Detection}. In the literature of robotic grasp detection, robots perceive and interact with their environment in one or more ways. For example, works like \cite{zheng2022human} used tactile sensing, while others mostly used computer vision. The robotic grippers used in these works also vary in a few ways. For example \cite{yao2023exploiting} solved the robotic grasping problem for a robotic hand of 5 fingers, \cite{liu2023hybrid} used a soft gripper with 3 fingers. Among methods that used learning-based grasp detection, \cite{zhang2020robotic} used classical machine learning techniques to perform robotic grasp detection. Some other works reviewed by \cite{mohammed2020review} based their work on deep reinforcement learning. In recent years, however, supervised robotic grasp detection found in deep learning has been the focus of many studies. These studies differ from each other regarding their configuration of grasping parameters. In this regard, \cite{jauhri2023learning} used 6-DOF grasp parameters. Furthermore, these deep learning-based studies can be classified by the type and modality of the dataset used. For instance, \cite{ainetter2021end} used RGB, and \cite{kumra2020antipodal} used RGB-D modalities. With the Cornell Grasping Dataset published by Lenz et. al., 2015\cite{lenz2015deep} and their 5-D grasp rectangle presentation for 4-DOF grasping, recent works have offered neural networks capable of predicting grasps in 2-dimensional settings more accurately and more efficiently. \cite{hosseini2020improving} directly regressed the feature maps into 5-D grasp parameters, while \cite{kumra2020antipodal} used a regression method to generate grasp maps. Some methods such as \cite{park2020single} combined object detection and grasp regression tasks by developing a multi-task network with two heads. Others such as \cite{asif2018graspnet} used a preceding segmentation stage in their network by introducing a new dataset of 42 different objects. Regarding representing the hand-object positioning information for grasp detection, Balazadeh et. al., 2023 \cite{balazadeh2023} proposed an approach-based dataset that used the nodes of the human fingertips to represent the approach information and label human-demonstrated grasps.

\textbf{Element Decomposition.} 
Decomposing an object into its main parts has been one of the main image processing and computer vision tasks. Works such as \cite{paschalidou2020learning} used hierarchical deep learning-based object decomposition methods using 3D point clouds. \cite{liu2018physical} decomposed tools and block towers into primitive components according to their geometry and physics. \cite{kawana2022unsupervised} decomposed various synthetic and real objects according to the mechanical joints in objects, for a functional part decomposition. Segmenting a scene into its main parts is a fundamental task in several fields such as medicine\cite{kazaj2022u}, autonomous vehicles\cite{chen2018importance}, and robotics\cite{asif2018graspnet}. In robotic grasp detection, segmentation has been the focal point of many recent studies. Lin et. al., 2019 \cite{lin2020using} used depth images to segment an object into its primitive components, and ranked grasps according to the generated point cloud. Asif et. al., 2018\cite{asif2018graspnet} proposed GraspNet, a grasp affordance segmentation network. Danielczuk, et. al., 2019 \cite{danielczuk2019segmenting} trained Mask R-CNN \cite{he2017mask} using synthetic data to segment unknown objects in depth images. Anitter et. al, 2022\cite{ainetter2021end} proposed a segmentation-based grasp detection network in challenging situations. Their method uses segmentation to refine the grasp predictions. MetaGraspNet \cite{chen2021metagraspnet_v0} was proposed to rank objects that are on top of each other in clutter scenes in different layers using a graph data structure.

In the current work, and as its main contribution, a method is proposed to detect the best grasping part of an object by providing an explicit representation of the approach of the gripper to an object and to find the best grasp point in the selected part. Built upon deep learning methods, the grasp detection model is trained on a new grasping dataset representing the approach of the gripper toward the object, which comes with element decomposition and grasp rectangle annotations. The element decomposition masks are used to train a Mask R-CNN to detect the different parts of an object, followed by a grasp detection head which utilizes the segmented parts and the gripper-object information to offer the most optimal grasp accordingly. The proposed end-to-end Approach-based Grasp Inference Learned from Element Decomposition (AGILE) method is shown in Fig \ref{fig:network}. The contributions of this paper are hereby:
\begin{enumerate}
    \item An approach-based robotic grasp detection dataset is proposed with element decomposition mask annotations, grasp rectangle annotations, and images of the gripper-object positioning relationship.
    \item A two-stage grasp inference method is presented for element decomposition and grasp prediction tasks. 
\end{enumerate}

The organization of this paper is as follows:
First, the proposed robotic grasping dataset is explained and the grasp rectangle and element decomposition annotating procedure is described. Next, the Approach-based Grasp Inference Learned from Element Decomposition (AGILE) is proposed which consists of two main stages, the element decomposition stage which is a Mask R-CNN network trained on the proposed dataset, and the grasp detection network responsible for processing the decomposed elements and approach information to detect the most optimal grasp accordingly. Afterward, the results of the trained networks in simulation and reality are comprehensively presented and analyzed. Lastly, the paper is concluded and future works are discussed.

\section{Approach-based Grasp Detection and Element Decomposition Dataset}
The approach-based grasp detection and element decomposition dataset is collected to explicitly represent the object and the approach of the gripper in images. The dataset is collected in Coppeliasim \cite{rohmer2013v} simulation environment. Fig. \ref{fig:appdata} shows the dataset collection procedure. Fig. \ref{fig:appdata}(a) top row shows the simulation environment. The scene includes two perspective vision sensors with a field-of-view of 60 degrees. Vision Sensor 1 views the scene from a top view, while the view of Vision Sensor 2 is an isometric configuration with a 45-degree orientation. Two dummies are placed in the environment to represent the approaching and grasping pose of the gripper, with both having 4-DOF to represent the DOF of a Delta parallel robot which will be used in experimental tests. The green dummies in Fig. \ref{fig:appdata} (a) are used to define different approaching and grasping poses by the human user. The images from the top view responsible for representing the object are recorded before the gripper has made an approach toward the object. Fig \ref{fig:appdata}(a) below row shows a top-view image of a mug in the top row. Next, the inverse kinematics is solved for the end-effector of the robot to reach the first set dummy, the approaching pose. The gripper used in Coppeliasim is the 2F-85 model by Robotiq\cite{Robotiq}, and the robotic arm that the gripper is mounted on is the Universal Robots UR10 robotic arm. The approach images are recorded when the gripper reaches the approach pose. Fig \ref{fig:appdata}(b) top row illustrates the environment when the gripper reaches the approaching pose, and Fig. \ref{fig:appdata}(b) below row shows the recorded image of the approach of the gripper. This image represents the gripper-object spatial information necessary to predict the best grasp of the selected part. The grasping pose is set based on the approach of the gripper and is determined by a dummy. The gripper moves from the approaching pose to the grasping pose and once the grasp is performed, the grasping parameters will be recorded. Fig \ref{fig:appdata}(c) top row shows a final grasp, for which the grasp rectangle is illustrated on the top-view image in Fig. \ref{fig:appdata}(c) below row, where red lines are the fingertips of the gripper, width of the rectangle is the final grasp width, and its center coordinates and orientation denote the final $x$, $y$, $z$, and $\theta$ of the gripper.
\begin{figure}[t]
\centerline{\includegraphics[ width = 0.5\textwidth]{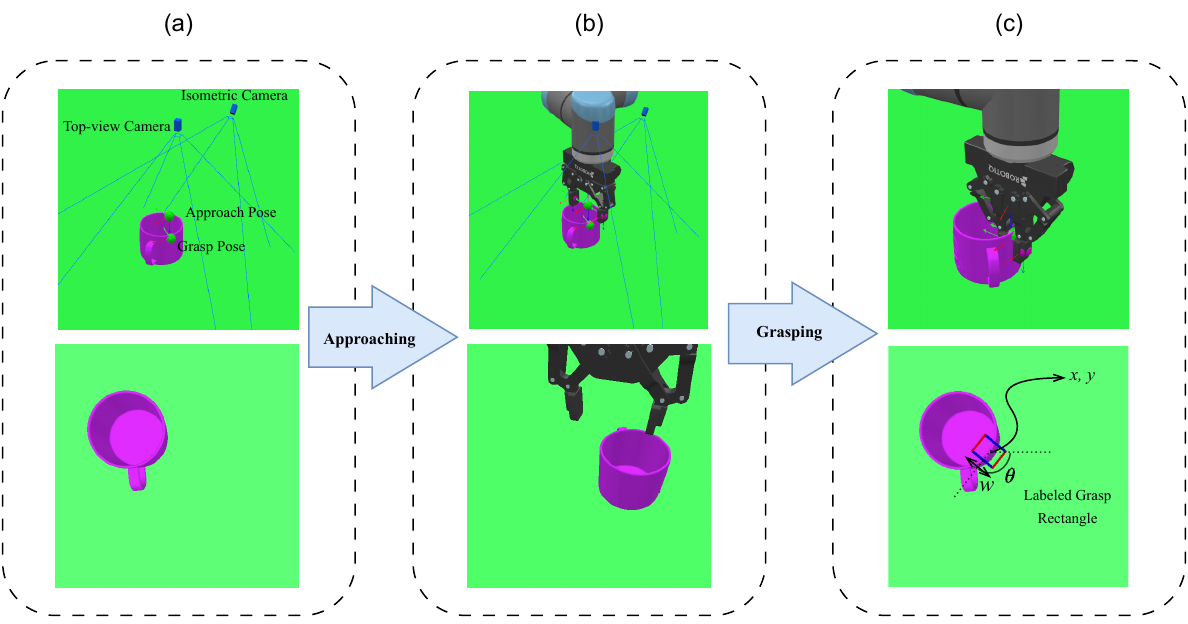}}
\caption{Collecting approach-based grasping dataset in the simulation environment. (a) shows how the vision sensors and the object are placed in the environment, without the approach having been made, with the recorded image of the object from the top-view camera in the below row. The top row of (b) shows the scene when the approach has been made, and the below row shows the recorded image of the approach of the gripper by the isometric camera shown in (a). The top row of (c) illustrates the gripper in the grasping pose, and the recorded grasp in the below row.}
\label{fig:appdata}
\end{figure}

Objects included in the grasping dataset training set are limited to 10 common household items such as mugs, bottles, padlocks, etc. Each object is placed in different locations and orientations, with different types of gripper-object positioning, whether leading to a new grasp part or not. Furthermore, each object appears in the dataset in different colors to the end of making the element decomposition and grasp detection models invariant to the color of the object. For each object available in the grasping dataset, one or more possible grasps are assumed, which means each object can be divided into one or more graspable elements, and in each element, one or more grasp poses can be offered. Different grasps with respect to different approaches for some objects are depicted in Fig. \ref{fig:samples}. 

\begin{figure}[t]
\centerline{\includegraphics[ width = 0.5\textwidth]{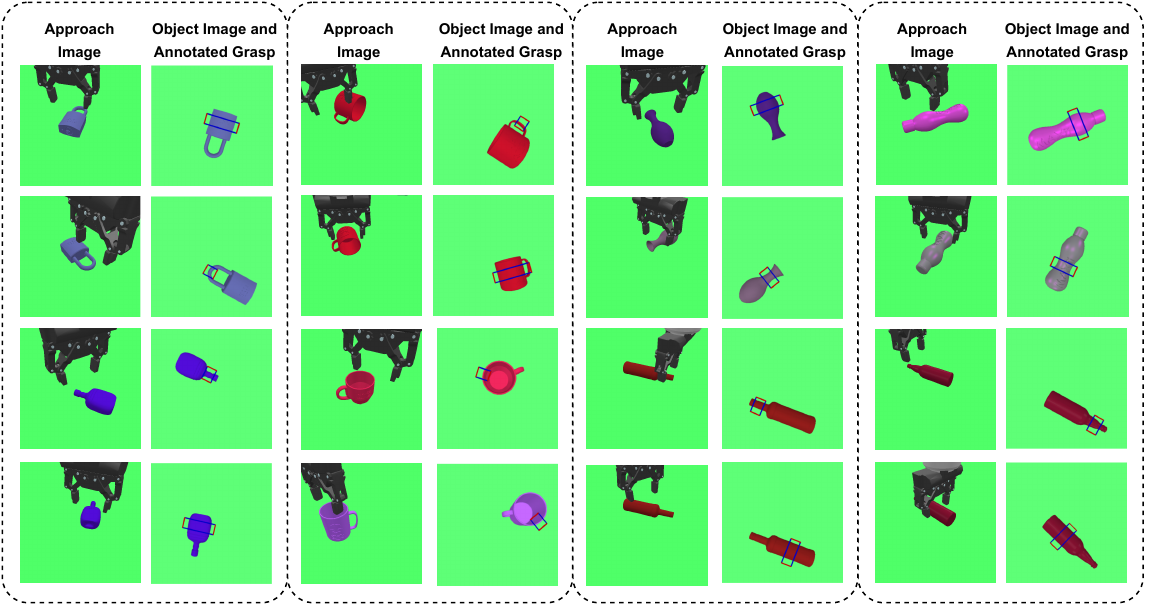}}
\caption{Samples from the training set of the proposed grasping dataset with grasp rectangle annotations. One can see that different approaches have led to different grasps of the objects.}
\label{fig:samples}
\end{figure}

\begin{figure}[t]
\centerline{\includegraphics[ width = 0.5\textwidth]{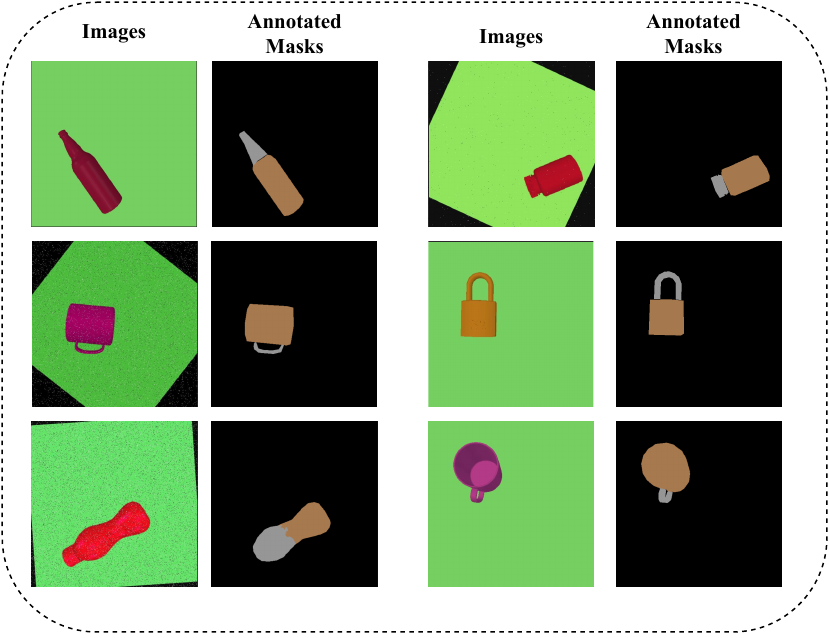}}
\caption{Samples of object images with augmentation transformations applied and their corresponding element decomposition masks.}
\label{fig:masks}
\end{figure}

\begin{figure*}[t]
\centerline{\includegraphics[ width = \textwidth]{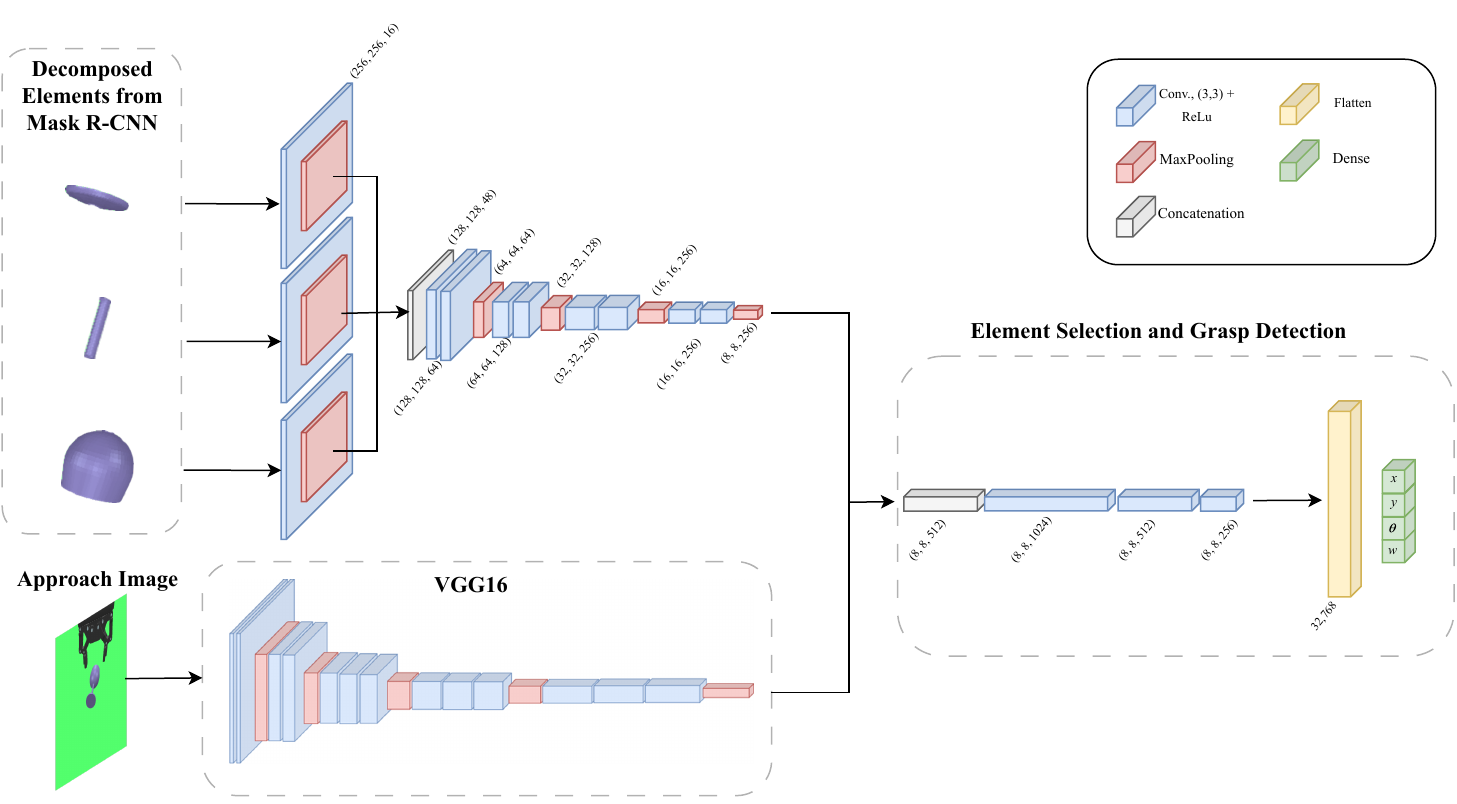}}
\caption{Grasp detection CNN. Decomposed by Mask R-CNN, the elements and the image representing the approach of the gripper are fed into the network to find the best grasping part. The network processes the feature maps to decide upon the most optimal decomposed element.}
\label{fig:gd_net}
\end{figure*}

\section{Element Decomposition and Approach-based Grasp Detection}
\label{section:ed_agile}

In order to train the element decomposition network, images of objects in the presented grasping dataset, which is recorded by the camera from the top view in the Coppeliasim environment, are annotated into different parts. The images are annotated using the Roboflow website\cite{Roboflow} ,which provides a fast and accurate segmentation annotation tool powered by the Segment Anything Model\cite{kirillov2023segment}. The dataset is annotated to directly address the needs in decomposing the different graspable elements in a single and integrated object, preventing this work from using other available datasets. Each segmentation annotation is bound with a class, denoting the name of the annotated area, e.g. ring, cylinder, sphere, etc. These classes will not be used in the grasp detection stage, but in the element decomposition stage and in order to evaluate the performance of the segmentation model. To enhance the generalization of the dataset, especially for bridging the simulation-to-reality gap, the dataset is augmented using the transformations provided by the Albumentations library\cite{buslaev2020albumentations}, which are rotating the object and the mask in the range of -45 to 45 degrees, Gaussian and ISO camera noise applied to the images, randomly altering the brightness and contrast of the images, randomly dropping out pixels of the images, and flipping the images either horizontally or vertically. Fig. \ref{fig:masks} illustrates samples of images and their corresponding annotations, some of which have been transformed by the augmentation configuration.

In order to split each object into its elemental graspable components, Mask R-CNN \cite{he2017mask} is trained on the annotated images of the objects in the grasping dataset. Due to its ROI-based approach, Mask R-CNN is capable of providing a suitable part segmentation performance. The trained Mask R-CNN network supplies the next CNN, the grasp detection network, with up to three segmented parts. It is notable that although open-source segmentation models such as SAM\cite{kirillov2023segment} are accessible and can provide accurate and robust segmentation in various scenes, the segmentation results of these networks do not accurately match the definition of element decomposition in robotic grasping, requiring this study to train a segmentation model from scratch using the collected and annotated dataset. Mask R-CNN is trained for 100 epochs with $1 \times 10^{-3}$ as the learning rate and SGD as the optimizer.

\begin{figure*}[t]
\centerline{\includegraphics[ width = \textwidth]{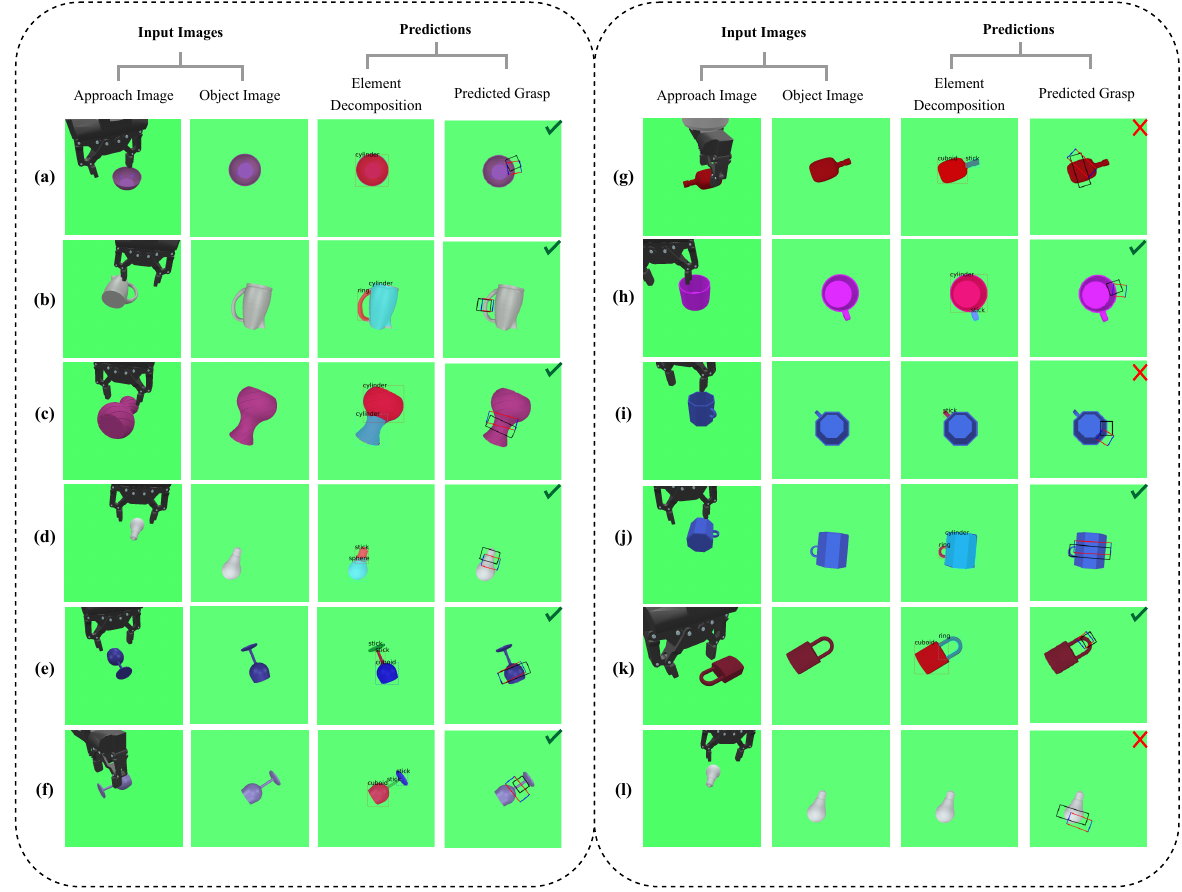}}
\caption{Grasp inference results for 12 unseen approach cases of seen and unseen objects. The predictions of the element decomposition and grasp prediction networks are reported separately. Check marks (\cmark) and cross marks (\xmark) denote successful and unsuccessful grasp predictions respectively.}
\label{fig:sim_tests}
\end{figure*}

The segmentation results of the Mask R-CNN will be applied to the object images from the grasping dataset to break up the object into different parts. The parts are subsequently provided to the grasp detection network, in addition to the approach image recorded by the isometric camera in the simulation environment. The grasp detection network is illustrated in Fig. \ref{fig:gd_net}. A convolution layer of size 16 and a max-pooling are applied to each segmented part, led by a concatenation layer to concat the three feature maps in depth. The pattern of two convolution layers and a max-pooling layer proceeds, resulting in feature maps with depth sizes of 64, 128, and 256. The approach image is supplied to a VGG16 \cite{simonyan2014very} pre-trained on the imagenet dataset. The final feature maps received from VGG16 are eventually input to a convolution layer with 256 filters. The feature maps given by the two networks are concatenated, leading to a feature map of depth 512. Three convolution layers are then applied to the concatenated feature maps. These layers are responsible for detecting the feature maps received from the part to which the gripper has made an approach, which is illustrated by the approach image. A flattened layer and a dense layer of 4 neurons are then applied to directly regress the grasping parameters, namely, $(x,y)$ of the grasp rectangle center, the angle, and the width of the grasp rectangle. The network is trained for 50 epochs, using a batch size of 64 and a learning rate of $1 \times 10^{-4}$, Adam as the optimizer, and mean absolute error as the loss function. Also, 20 percent of the training set is used as validation data in order to evaluate the training performance, making a training and validation portion of 944 images and 236 samples respectively. In order to address the simulation-to-reality domain adaptation for the grasp detection network, ISO, Gaussian, and Multiplicative noise models available by Albumentations are applied along with a color transformation that addresses the appearance of the white gripper, in contrast to the similar black model in the Coppeliasim environment.

\section{Results and Experiments}
In this section, the element segmentation and grasp detection performance of the trained models are presented in both simulation and real-world scenarios. A test set in the simulation is annotated to evaluate the grasp success rate accordingly and by using the conventional evaluation metrics. In reality, grasps are performed and their success is assessed by the physical grasping criteria.

\subsection{Element Segmentation and Grasp Detection Results}

\begin{table}[t]
    \caption{Element Decomposition results for different Min Detection Confidence (MDC) }
    \centering
    \begin{tabular}{|c|c|c|c|}
    \hline
        \textbf{Element} & \textbf{Dice(MDC=0.9)} & \textbf{Dice(MDC=0.85)} & \textbf{Dice(MDC=0.8)} \\ \hline
        \textbf{Cuboid} & 90.0 & 90.8 & 90.8 \\ \hline
        \textbf{Sphere} & 91.3 & 92.9 & 92.9 \\ \hline
        \textbf{Cylinder} & 94.1 & 94.0 & 93.9 \\ \hline
        \textbf{Ring} & 86.7 & 86.7 & 86.7 \\ \hline
        \textbf{Stick} & 87.6 & 86.6 & 85.6 \\ \hline\hline
        \textbf{Mean} & 90.0 & 90.2 & 90.0 \\ \hline
    \end{tabular}
    \label{tab:dice}
\end{table}
As discussed in Section \ref{section:ed_agile}, the first stage of the inference pipeline is Mask R-CNN, responsible for element decomposition. In order to evaluate the performance of Mask R-CNN, the Dice Similarity Coefficient ($DSC$) is used, which can be formulated as follows:
\begin{align*}
    DSC =2 \frac{A \cap B}{A + B} 
\end{align*}
where $A$ is the prediction mask and $B$ is the ground truth mask.
Detailed element decomposition results for different element classes and different values of Min Detection Confidence (MDC) are depicted in Table \ref{tab:dice}. According to the mean values for each MDC, the best element decomposition results are derived with an MDC of 0.85, which will be used during the grasp inference stage.

To assess the trained network on a wide range of object complexity and approach complexity, 100 unseen approach and grasping cases are recorded. Among these images, 50 images are of the objects that appeared in the training set of the proposed grasping dataset, although with an unseen approach, and others are images with objects not seen by the model with different levels of complexity. This paper uses the 3-DOF grasp success criteria commonly used in literature. According to the criteria, a grasp is successful if the difference between the ground truth and the predicted grasp angles is lower than 30 degrees, and if the Jaccard index between the ground truth and the predicted grasp rectangles is more than 25 percent. The Jaccard index ($J$) is defined as follows:
\begin{align*}
    J &=\frac{g \cap \hat{g}}{g \cup \hat{g}} 
\end{align*}
where $g$ and $\hat g$ denote the ground truth and predicted grasp rectangles respectively. 

Fig. \ref{fig:sim_tests} shows 12 grasp predictions for seen and unseen objects in the test set, in which rows (g), (i), and (l) illustrate unsuccessful grasping cases while others are cases of successful grasp prediction. In this figure ground truth, grasp rectangles are drawn with solid black color and predicted grasp rectangles with red and blue colors. All these cases include grasps and approach information not available in the training set. Rows (i) and (l) in Fig. \ref{fig:sim_tests} show unsuccessful element decomposition results for unseen objects which have led to unsuccessful grasping results. In row (i), the Mask R-CNN network has detected only one element and has missed out on the cylindrical part of the hexagonal mug, the body, and although the predicted grasp is a generally successful grasp, it does not accurately follow the approach information. Similarly in row (l), the trained Mask R-CNN has failed to decompose any element. Contrary to row (l), row (d) shows a grasping case of the same object, a bulb lamp, whose elements have been decomposed successfully, followed by a successfully detected grasp. These cases point out the importance of decomposing the elements for generalization on unseen objects. Rows (g), (h), and (k) depict element segmentation and grasp detection performance on three seen objects, a cuboid bottle, a mug, and a padlock. All of these cases but row (g) show successful grasp predictions. Although involving a successful element decomposition stage, row (g) is most possibly due to the positioning of the gripper in the approach image. The same goes for row (f), where the successfully predicted grasp has not yielded a great Jaccard index due to the limited representation of the gripper-object information in the approach image. Row (e) shows a case of the same object as row (f), with a successfully predicted grasp with a high Jaccard index, thanks to the thorough and accurate element segmentation and approach information presented to the grasp detection network. Row (c) also shows a case of a successful grasp for a large vase with a lower number of Jaccard index, where part of the object is blocked by its bigger part in its corresponding approach image captured from only a single point-of-view isometric camera. These cases show using one approach image from a single Point-of-view may not lead to very accurate gripper-object feature extraction in all scenarios. 

Table \ref{table:sim_results} shows the grasp detection performance over the labeled test set of 100 grasping cases and 12 objects in the Coppeliasim simulation, all of which include approaches and grasps not seen by the grasp detection network during training. Among the 12 objects in the experiments, 4 objects were seen during the training and 8 objects are unseen objects. Each seen and unseen object portion has 50 grasping cases. The objects appear in different sizes and colors. Table \ref{table:sim_results} shows overall grasp success rate results over the test set and Table \ref{table:obj_results} shows grasping results for each object separately. 

\begin{table}[t]
\caption{Grasp success rate and mean Jaccard index for seen and unseen objects.}
\centering
\label{table:sim_results}
\begin{tabular}{|c|c|c||c|}
\hline
\textbf{Objects} & \textbf{Seen Objects} & \textbf{Unseen Objects} & \textbf{Overall} \\ \hline
\textbf{Grasp Success Rate} & 90 \% & 78 \% & 84 \% \\ \hline
\textbf{Mean Jaccard Index} & 66.32 \% & 45.63 \% & 55.98 \% \\ \hline
\end{tabular}
\end{table}

\begin{table}[t]
\caption{Grasp success rate for each object.}
\centering
\label{table:obj_results}
\begin{tabular}{|c|c|c|}
\hline
\textbf{Objects Type} & \textbf{Objects} & \textbf{Grasp Results} \\ \hline \hline
\multirow{4}{*}{\textit{\textbf{Seen Objects}}} & \textit{Mug} & 11/13 \\ \cline{2-3} 
 & \textit{Beer Bottle} & 12/13 \\ \cline{2-3} 
 & \textit{Cubic Bottle} & 12/12 \\ \cline{2-3} 
 & \textit{Padlock} & 10/12 \\ \hline
\multirow{8}{*}{\textit{\textbf{Unseen Objects}}} & \textit{Drink Bottle} & 7/10 \\ \cline{2-3} 
 & \textit{Bowl} & 4/4 \\ \cline{2-3} 
 & \textit{Hexagonal Cup} & 7/8 \\ \cline{2-3} 
 & \textit{Large Vase} & 4/6 \\ \cline{2-3} 
 & \textit{Electric Kettle} & 4/4 \\ \cline{2-3} 
 & \textit{Banana} & 5/6 \\ \cline{2-3} 
 & \textit{Goblet} & 6/8 \\ \cline{2-3} 
 & \textit{Bulb Lamp }& 2/4 \\ \hline
\end{tabular}
\end{table}

\begin{table}[t]
\caption{Grasp success rate for different thresholds that yield a successful grasp. All images are unique and not used in training, although seen and unseen objects are included.}
\centering
\label{table:jaccard}
\begin{tabular}{|c|c|c|c|c|}
\hline
\textbf{Jaccard Threshold  }        &\textbf{ 20\%}  &\textbf{ 25\%} & \textbf{30\%} & \textbf{35\%} \\ \hline
\textbf{Seen Objects Grasp Success}   & 94\% & 90\% & 86\% & 80\%\\ \hline
\textbf{Unseen Objects Grasp Success} & 86\% & 78\% &  68\%  & 62\%  \\ \hline \hline
\textbf{Overall} & 90\% & 84\% &  77\%  & 71\%  \\ \hline
\end{tabular}
\end{table}

Regarding the grasp rectangle criteria for grasp evaluation, different Jaccard index values can be chosen as the grasp success threshold. Commonly set to 25 percent in literature \cite{redmon2015real}, the threshold decides what the minimum Jaccard index for a grasp prediction is to be successful. Table \ref{table:jaccard} shows the grasp success rate on seen and unseen objects according to different Jaccard index thresholds.

\subsection{Experimental Results}
\begin{figure}[t]
\centerline{\includegraphics[ width = 0.45\textwidth]{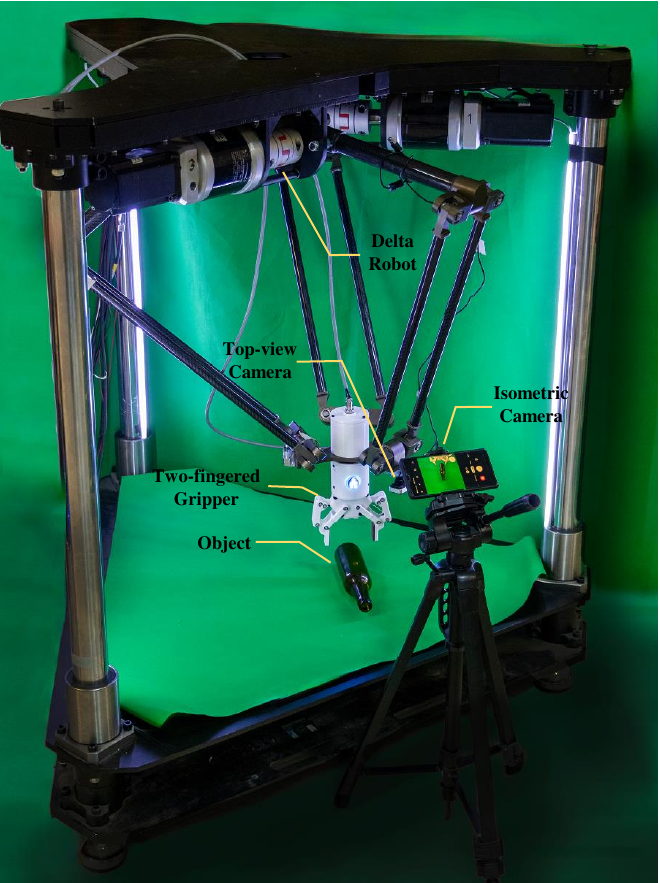}}
\caption{The practical setup and its components. The gripper has approached the object and the isometric camera records the approach information.}
\label{fig:setup}
\end{figure}

\begin{figure}[t]
\centerline{\includegraphics[ width = 0.5\textwidth]{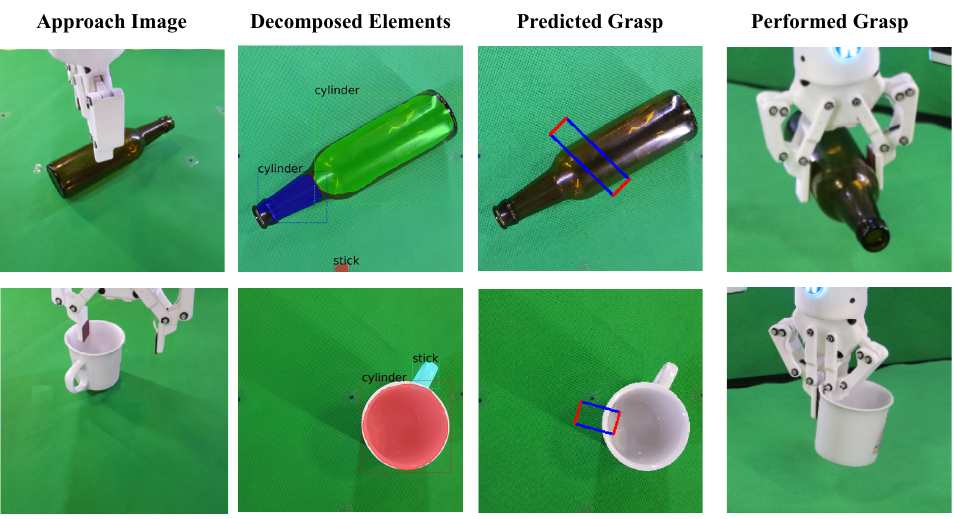}}
\caption{Two cases of physical grasp experiments. The approach image, element decomposition results, grasp prediction results, and the final performed grasps are depicted.}
\label{fig:real_tests}
\end{figure}

In order to evaluate the grasping performance of the network, the experimental setup in Fig. \ref{fig:setup} is used. This setup includes a 2-fingered gripper by \cite{navidhand2023} mounted on a 3-DOF Delta parallel robot. The end-effector of the robot is equipped with a rotary component, which provides the rotation of the end-effector, $\theta$. The front camera of a Samsung Galaxy Note 8 mobile phone is used as the isometric camera, providing the gripper-object approach information, and ODROIDUSB-CAM 720PHD is used as the top-view camera. Practical grasping experiments presented in this section are physically performed by the gripper, meaning the grasp success in these experiments is evaluated qualitatively and by visual inspection. 

Physical grasp performance results are reported in Table \ref{table:real_tests} for 4 different objects, two available in the synthetic training dataset and two unseen. Two cases of successfully performed grasps along with the predicted decomposed elements and predicted grasp rectangle are depicted in Fig. \ref{fig:real_tests}. Practical experiments have yielded a 70\% overall grasp success rate. 

\begin{table}[t]
\centering
\caption{Grasp inference results in reality environment.}
\label{table:real_tests}
\begin{tabular}{|c|c|}
\hline
\textbf{Object} & \textbf{Grasp Results} \\ \hline\hline
\textbf{Mug} & 3/5 \\ \hline
\textbf{Beer Bottle} & 4/5 \\ \hline
\textbf{Banana} & 4/5 \\ \hline
\textbf{Bowl} & 3/5 \\ \hline
\end{tabular}
\end{table}

\section{Conclusion}
This work presented a novel grasp detection method based on convolutional neural networks by representing the gripper-object spatial information and decomposing objects into distinct elements. A new robotic grasping dataset is proposed with element decomposition and grasp rectangle annotations in the Coppeliasim simulation environment. The method obtained an 84\% grasp success rate on 100 unseen approaching and grasping cases of both seen and unseen objects. Lastly, by augmenting the simulation training set using transformations which can account for the differences between reality and simulation scenarios, the grasp detection performance of the proposed CNN is evaluated in physical grasping scenarios, which is 70\% for 20 grasping cases. Additionally, the 78\% grasp success rate on unseen objects in simulation-based experiments revealed that although the training dataset involved approaching and grasping cases of only 10 objects, the proposed element decomposition stage has assisted the network in generalization over unseen objects. Early experiments with an end-to-end grasp detection network with the same input and output configuration and without an element decomposition stage did not yield results as accurate and stable. Furthermore, the failed grasp inference cases indicated that representing the approach of the gripper by one isometric vision sensor brings about limits in terms of the hand-object feature extraction such as the object getting blocked by the gripper or a limited representation of the gripper positioning. These limits can lead to a grasp prediction not complying with the approaching pose. Using two isometric cameras to record the scene when the gripper is in the approaching pose can help overcome the discussed limitations. Also, by changing the modality of the isometric cameras from RGB to RGB-D or Depth, a more comprehensive representation of the gripper-object information is possible. In order to improve the grasp success results of unseen objects, the training set can expand to more objects to enable the trained model for better generalization. Moreover, the gap between the grasp prediction results in reality and simulation points out the shortcomings of the applied augmentation transformations on the simulation-based training dataset to address the sim-to-real challenges. For a more effective supervised sim-to-real domain adaptation, a supplementary dataset should be collected in the reality environment as future work. In addition and as an ongoing work, the authors will add another intermediary stage to the proposed grasp inference pipeline after the element decomposition stage to select the graspable decomposed part before detecting the final grasping pose. The new grasp inference pipeline can make the grasp detection task more efficient and more accurate.

\bibliographystyle{IEEEtran}
\bibliography{cas-refs}

\end{document}